\documentclass[a4paper,fleqn]{cas-sc}

\usepackage{microtype}
\usepackage{enumitem}
\usepackage{placeins}
\usepackage[authoryear]{natbib}
\usepackage{bbm}

\hypersetup{colorlinks=true, linkcolor=blue!55!black, citecolor=blue!55!black, urlcolor=blue!55!black}

\ExplSyntaxOn
\cs_set:Npn \__first_footerline: {}
\ExplSyntaxOff

\newcommand{\R}{\mathbb{R}}

\newcommand{\auroc}{\mathrm{AUROC}}
\newcommand{\auprc}{\mathrm{AUPRC}}
\newcommand{\fpr}{\mathrm{FPR}}

\begin{document}
\let\WriteBookmarks\relax
\def\floatpagepagefraction{1}
\def\textpagefraction{.001}

\shorttitle{OOD-RL-Bench}
\shortauthors{Mittag et~al.}

\title[mode=title]{OOD-RL-Bench: A Benchmark Framework for Out-of-Distribution Detection in Reinforcement Learning}

\author[1]{Emil Mittag}
\cormark[1]
\ead{emil.mittag@research.deakin.edu.au}
\author[1]{Richard Dazeley}
\ead{richard.dazeley@deakin.edu.au}
\author[2]{Peter Vamplew}
\ead{p.vamplew@federation.edu.au}

\affiliation[1]{
  organization={Deakin University},
  city={Geelong},
  state={Victoria},
  postcode={3220},
  country={Australia}
}
\cortext[1]{Corresponding author}

\affiliation[2]{
  organization={Federation University Australia},
  city={Ballarat},
  state={Victoria},
  postcode={3350},
  country={Australia}
}

\begin{abstract}
Reliable reinforcement learning (RL) agents must maintain operational integrity amidst sensor malfunctions, dynamic disturbances, and slow environmental shifts. The detection of out-of-distribution conditions is pivotal to determining when an agent's observations, transitions, or trajectory dynamics deviate from the assumptions underpinning its policy training. Current out-of-distribution (OOD) detection benchmarks typically evaluate image classifiers or static low-dimensional datasets, failing to account for the complex, action-dependent temporal structure inherent in RL trajectories. To address this gap, we present \emph{OOD-RL-Bench}, a comprehensive and extensible framework designed to evaluate OOD detectors against categories of anomalies injected into RL trajectories. Detectors and anomaly injectors are integrated through shared interfaces and configuration, which allows new scoring methods and perturbation families to be evaluated without modification of the core benchmark loop. We evaluate the utility of the framework using a Deep Q-Network policy within the LunarLander-v3 environment. We assess the performance of each detector across a suite of anomaly types using matched-time AUROC, matched-time AUPRC, matched-time false-positive rate, detection delay, and segmented-onset metrics. Our analysis reveals significant performance variance across anomaly types: observation perturbations and regime switches are identified with high accuracy by several methods, while observation delay and action-conditioned dynamics remain difficult even when post-onset anomaly scores are compared against clean scores from the same timesteps. We make the framework, trained policy checkpoint, and complete results publicly available as a reproducible artefact.
\end{abstract}

\begin{keywords}
out-of-distribution detection \sep reinforcement learning \sep anomaly detection \sep benchmark \sep trajectory analysis
\end{keywords}

\maketitle
\thispagestyle{plain}
\pagestyle{plain}
\hypersetup{
  pdfsubject={Benchmarking out-of-distribution detection in reinforcement learning},
  pdfkeywords={out-of-distribution detection, reinforcement learning, anomaly detection, benchmark, trajectory analysis}
}
\section{Introduction}
\label{sec:introduction}

Reinforcement learning (RL) agents are increasingly utilised in settings where runtime conditions frequently diverge from those encountered during training, spanning applications such as robotics, autonomous driving, and industrial control. Because a trained policy provides no assurance of appropriate behaviour when operating outside its training distribution, an unobserved shift in observations, dynamics, or actuation can lead to silent performance degradation or unsafe outcomes within safety-critical systems. Consequently, the reliable detection of out-of-distribution (OOD) conditions is essential for the trustworthy deployment of RL: it enables an agent to recognise when its competence assumptions are no longer valid and to defer, intervene, or revert to a safe controller.

Detecting such shifts in RL is, however, not a straightforward application of existing OOD methods. RL agents select actions within sequential decision-making environments and learn from the outcomes of those actions \citep{sutton2018}, which renders OOD detection in RL structurally distinct from supervised image classification. In classifier benchmarks, the test instance is typically an externally provided input. Conversely, within an RL deployment, the policy actively contributes to generating its future input distribution, as every action alters the subsequent state, the reward sequence, and the state visitation distribution. Consequently, a detector for RL trajectories must be evaluated not merely on its capacity to identify unusual observations, but also on its ability to remain effective when anomalies manifest through transition dynamics, action-conditioned prediction errors, temporal drift, or delayed observations.

General OOD benchmarks have significantly enhanced reproducibility within supervised environments. For instance, OpenOOD standardises broad OOD evaluation protocols for image-classification contexts and has subsequently extended the framework to encompass large-scale data and foundation-model environments \citep{yang2022,zhang2023}. Time-series anomaly-detection benchmarks have similarly broadened the scale and rigour of temporal anomaly assessment, incorporating large curated datasets and unified evaluation pipelines \citep{paparrizos2022,qiu2025}. While these resources are significant, they do not entirely capture the closed-loop, action-conditioned structure characteristic of RL trajectories. \citet{mueller2022} contended that anomaly detection in RL warrants distinct consideration because sequential decision-making alters the interpretation of anomalous data. 

RL-specific literature on OOD and anomaly detection has commenced addressing this deficiency. \citet{mohammed2021} proposed an RL-OOD benchmark grounded in physical-parameter modifications and observation corruptions. \citet{danesh2021} introduced OOD dynamics benchmarks derived from RL environments and assessed recurrent prediction-error methods. \citet{haider2023} proposed the use of probabilistic dynamics models and bootstrapped ensembles for semantic perturbations within Markov decision processes. More recent research, including DEXTER (Detection via Extraction of Time Series Representations), has further categorised RL-OOD into distinct classes such as sensory and semantic anomaly scenarios \citep{nasvytis2024}. These contributions demonstrate that the field is no longer devoid of resources; the contribution of OOD-RL-Bench consequently lies not in being the inaugural RL OOD benchmark, but in offering a controlled trajectory-scoring protocol for comparing a restricted set of interpretable detectors across post-hoc observation, temporal, and distributional perturbations and online action/dynamics perturbations.

\paragraph{Contributions.} This work presents: (i) an extensible RL-OOD benchmark framework, configured here with eight detectors and nine anomaly injectors, that is designed so additional detectors and injectors can be registered through the same interfaces; (ii) a detector taxonomy categorising marginal-state, temporal-feature, transition-model, reconstruction, ensemble-uncertainty, sequential-change, non-parametric-density, and nonlinear-dynamics signals; (iii) a mixed anomaly protocol that evaluates observation, temporal, and distributional anomalies post hoc while evaluating action and dynamics anomalies through live environment rollouts; and (iv) a reporting template distinguishing between matched-time threshold-free discrimination, thresholded decisions, temporal delay, null-baseline score drift, and operating-point false-positive behaviour.

\paragraph{Paper organisation.} The manuscript adopts a structure that progresses from motivations and contracts to reusable framework components, culminating in a singular concrete evaluation. Section~\ref{sec:related} situates the benchmark within the context of supervised OOD detection, time-series anomaly detection, and existing literature specific to RL-based OOD approaches. Section~\ref{sec:problem} defines the settings for trajectories, detectors, and anomaly injectors. Section~\ref{sec:framework-design} illustrates the end-to-end framework map, followed by Sections~\ref{sec:detectors} and~\ref{sec:injection}, which detail the extensible interfaces for detectors and anomaly injectors. The evaluation methodology is outlined in Section~\ref{sec:evaluation}, covering metrics, experimental design, anticipated detector behaviour, and the rollout protocol; this is followed by Section~\ref{sec:setup}, which instantiates these abstractions for the LunarLander-v3 environment. Section~\ref{sec:results} presents the empirical results, comparing expected versus observed detector performance, while Section~\ref{sec:discussion} revisits architecture, extensibility, reporting standards, caveats, and limitations. This structural arrangement allows readers to grasp the benchmark's reusable contracts, understand the specific experimental configuration, and observe what the framework demonstrates in practice.

\section{Related Work}
\label{sec:related}

\paragraph{Static OOD detection.} Early neural OOD detection research demonstrated that neural classifiers can assign high confidence to anomalous inputs, prompting the development of straightforward confidence baselines for identifying such instances \citep{hendrycks2017}. Feature-space Mahalanobis scoring subsequently employed class-conditional Gaussian discriminant assumptions to identify OOD and adversarial inputs within neural representations \citep{lee2018}. The OpenOOD framework consolidates this body of work by standardising the evaluation of OOD, open-set recognition, and anomaly detection under a generalised OOD framework, primarily within image-classification settings \citep{yang2022,zhang2023}. While these methods are pertinent to state-space scoring, they are inadequate as comprehensive RL benchmarks because they do not account for actions, rewards, or transition structures.

\paragraph{Time-series anomaly detection.} The detection of anomalies in sequential data possesses a substantial statistical history. The cumulative-sum procedure, introduced by \citet{page1954}, remains a fundamental method for detecting sustained shifts in the mean. Broader treatments of change detection formalise the identification of abrupt changes through likelihood and sequential testing perspectives \citep{basseville1993}. Contemporary time-series anomaly benchmarks facilitate larger and more reproducible comparisons across statistical, machine learning, and deep learning detectors \citep{paparrizos2022,qiu2025}. Time-series methods are relevant to RL given the temporal nature of trajectories; however, passive time-series benchmarks typically do not condition on the agent's actions or incorporate reward and transition labels derived from an MDP.

\paragraph{RL-specific OOD and anomaly detection.} The most directly related literature encompasses RL-OOD and RL anomaly detection. These intersecting fields concentrate on delineating distributional shifts, anomalous dynamics, and runtime perturbations within agent–environment interactions. RL-OOD detection typically frames the problem as identifying states, transitions, observations, rewards, or trajectories that diverge from the agent's training distribution. Conversely, RL anomaly detection maintains a broader scope, focusing on anomalous execution events, faults, or perturbations that may compromise agent behaviour. \citet{mohammed2021} alter physical parameters within non-visual environments and perturb observations within visual environments to assess OOD methods for deep RL. \citet{danesh2021} concentrate on the detection of OOD dynamics and present RL-relevant benchmarks incorporating prediction-error baselines. \citet{haider2023} identify severe MDP perturbations utilising probabilistic dynamics models and bootstrapped ensembles. \citet{nasvytis2024} differentiate sensory and semantic RL-OOD settings and introduce DEXTER as a detector that extracts time-series features from environment observations and applies an isolation-forest ensemble. \citet{prashant2025} conceptualise transition-level OOD execution using learned transition models and conformal-style guarantees. OOD-RL-Bench is situated within this body of work as a controlled, detector-agnostic trajectory perturbation benchmark: post-hoc perturbation is retained where it is faithful to the anomaly channel, while live rollouts are used where action or transition faults would otherwise be misrepresented.

\paragraph{Model-based RL and uncertainty.} Probabilistic ensemble dynamics models are frequently driven by model-based RL and uncertainty-aware prediction. PETS integrates probabilistic neural dynamics models with ensembles and trajectory sampling to propagate uncertainty within model-based control \citep{chua2018}. Deep ensembles offer a straightforward and scalable avenue to predictive uncertainty, potentially demonstrating higher uncertainty on OOD inputs in supervised settings \citep{lakshminarayanan2017}. In RL-OOD detection, these concepts underpin both \texttt{pedm}, which employs sampled residual error under an action-conditioned transition model, and \texttt{ensemble disagreement}, which utilises disagreement among bootstrapped models as a signal of epistemic-uncertainty.

\paragraph{Reconstruction, density, features, and nonlinear dynamics.} Autoencoders are frequently employed for anomaly detection through training on nominal data, where substantial reconstruction errors are flagged as anomalies; \citet{sakurada2014} offer a pioneering application of neural autoencoders for this purpose, utilising nonlinear dimensionality reduction. Kernel density estimation is rooted in the non-parametric density-estimation research of \citet{parzen1962}; within an OOD detection framework, low estimated density regarding the in-distribution state or transition distribution constitutes the anomaly score. Automated time-series feature extraction demonstrates that extensive collections of interpretable temporal statistics can be utilised to characterise time-series structure \citep{fulcher2018}. Lyapunov-exponent estimation derived from time series provides a robust method for quantifying local divergence within nonlinear dynamical systems \citep{wolf1985,kantz2004}; this concept informs the development of the \texttt{lyapunov divergence} detector for anomalies manifesting as alterations in local trajectory divergence.

\section{Problem Setting}
\label{sec:problem}

Let a trajectory be a fixed-length sequence
\begin{equation}
\tau = \{(s_t, a_t, r_t, s_{t+1}, d_t)\}_{t=0}^{T-1},
\end{equation}
where $s_t \in \R^{d_s}$ is the observed state, $a_t \in \mathcal{A}$ is the action, $r_t \in \R$ is the reward, $s_{t+1}$ is the next state, and $d_t$ indicates termination. A detector $D$ is fitted using in-distribution trajectories $\mathcal{T}_{\mathrm{ID}}$ and returns an anomaly-score sequence
\begin{equation}
D(\tau) = (z_0, z_1, \ldots, z_{T-1}),
\end{equation}
where larger $z_t$ means greater evidence of an anomaly at timestep $t$.

The benchmark presupposes an anomaly injector, $I_j$, which yields a perturbed trajectory by modifying a stored rollout post-collection or by perturbing the live environment during the collection phase. In both instances, detectors are fitted prior to evaluation and are subsequently provided with only the resulting trajectory record, denying access to the simulator. The post-hoc mode retains a byte-identical clean rollout for observation, temporal, and distributional perturbations; the online mode is utilised for action and dynamics perturbations, as the state visitation distribution is contingent upon the manner in which the policy interacts with the fault.

The concrete environment and trained policy used for the demonstration experiment are introduced after the general benchmark methodology in Section~\ref{sec:setup}.

\section{Framework Design}
\label{sec:framework-design}

\begin{figure}[pos=htbp]
\centering
\includegraphics[width=\linewidth]{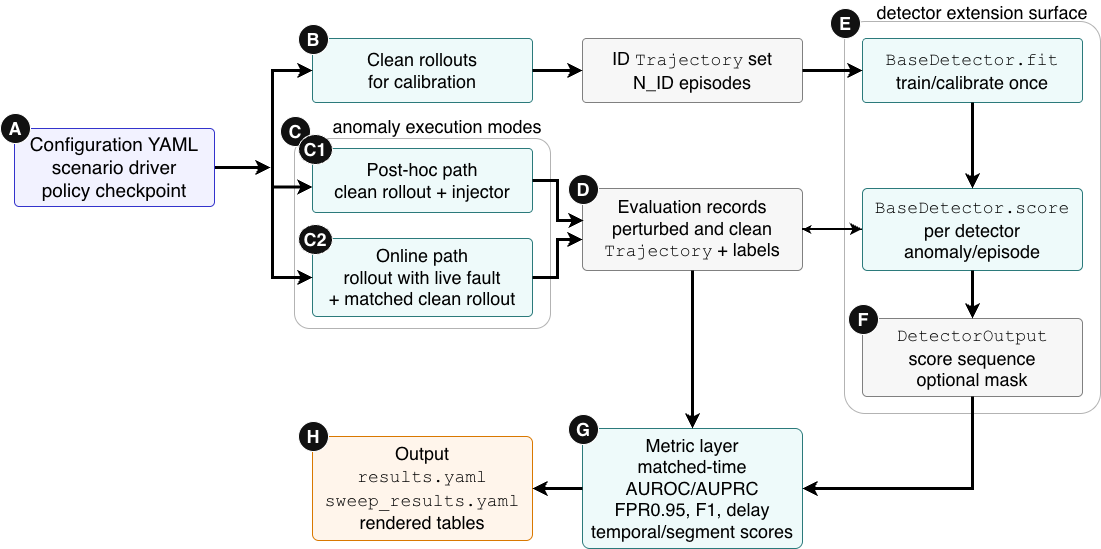}
\caption{OOD-RL-Bench framework architecture. Annotation labels correspond to (A) configuration, scenario, and policy input; (B) clean calibration rollouts and in-distribution trajectories; (C) anomaly execution modes, divided into (C1) post-hoc and (C2) online pathways; (D) clean and perturbed evaluation records; (E) detector fitting and scoring; (F) DetectorOutput records; (G) metric aggregation; and (H) rendered outputs and reproducibility artefacts}
\label{fig:framework-design}
\end{figure}

Figure~\ref{fig:framework-design} illustrates the high-level architecture employed throughout the remainder of this manuscript. The configuration, scenario driver, and policy checkpoint (Figure~\ref{fig:framework-design}A) initialise the benchmark run and feed the clean calibration rollouts (Figure~\ref{fig:framework-design}B), which generate the in-distribution trajectories that are used to fit the detectors. Anomaly execution can then proceed via either post-hoc injection into a stored rollout (Figure~\ref{fig:framework-design}C1) or through online perturbation during the interaction of a matching clean rollout with the environment (Figure~\ref{fig:framework-design}C2). These anomaly-execution pathways function as extension points, permitting the registration of additional post-hoc or online injectors while maintaining the same trajectory output contract. Both paths produce clean and perturbed evaluation records (Figure~\ref{fig:framework-design}D), which are scored by the detector (Figure~\ref{fig:framework-design}E) to generate DetectorOutputs (Figure~\ref{fig:framework-design}F). The detector similarly serves as an extension point; new detectors can be incorporated, provided they adhere to the requisite fit-and-score contract. Finally, the metric layer (Figure~\ref{fig:framework-design}G) aggregates these records to form reproducibility artefacts and rendered outputs (Figure~\ref{fig:framework-design}H).

\FloatBarrier

\section{Detector Suite}
\label{sec:detectors}

By default, the configuration evaluates the eight baseline detectors listed in Table~\ref{tab:detectors}. These detectors are reference implementations: the framework is engineered to allow for the addition and removal of any number of user-defined detectors through the detector interface and configuration file.

\begin{table}[t]
\centering
\small
\caption{Detector families used in the OOD-RL-Bench framework.}
\label{tab:detectors}
\begin{tabular}{p{0.25\linewidth}p{0.29\linewidth}p{0.35\linewidth}}
\toprule
Detector & Primary signal & Main limitation \\
\midrule
\texttt{mahalanobis distance} & Distance from fitted ID mean under covariance scaling & Sensitive to covariance estimation and approximately elliptical assumptions \\
\texttt{time series features} & Window-level temporal descriptors & Depends on chosen features and window length \\
\texttt{pedm} & Minimum sampled action-conditioned transition error & Requires reliable dynamics fitting \\
\texttt{autoencoder} & Reconstruction error after training on ID data & Can reconstruct some OOD inputs too well \\
\texttt{ensemble disagreement} & Variance among bootstrapped predictive models & Disagreement is not a calibrated probability by itself \\
\texttt{cusum} & Accumulated deviation from an ID reference statistic & Best suited to sustained shifts, not isolated anomalies \\
\texttt{kde density} & Negative log non-parametric density estimate & Degrades with dimension and bandwidth mis-specification \\
\texttt{lyapunov divergence} & Local trajectory divergence in reconstructed phase space & Sensitive to embedding, noise, and finite-sample effects \\
\bottomrule
\end{tabular}
\end{table}

\subsection{Mahalanobis distance}

Mahalanobis distance constitutes a classical covariance-normalised distance measure \citep{mahalanobis1936}. The Mahalanobis detector fits an in-distribution mean vector $\mu$ and covariance matrix $\Sigma$ across selected trajectory features $x_t$ (e.g., $s_t$, $(s_t,a_t)$ embeddings, or transition vectors $(s_t,a_t,s_{t+1})$). Its score is
\begin{equation}
 z_t = (x_t - \mu)^\top (\Sigma + \lambda I)^{-1}(x_t - \mu),
\end{equation}
where $\lambda I$ serves as a ridge regularisation term employed when the empirical covariance is ill-conditioned. In the context of machine-learning OOD detection, Mahalanobis feature scores are frequently linked to the Gaussian-discriminant approach of \citet{lee2018}. Within this benchmark, the detector functions as a transparent state or transition outlier detector, rather than a deep classifier feature detector.

\subsection{Time series features}

The time-series feature detector transforms a rolling window $W_t = \{x_{t-L+1},\ldots,x_t\}$ into a vector of descriptive statistics. Suitable descriptors encompass means, variances, slopes, autocorrelations, ranges, spectral summaries, and action-conditioned residual summaries. The detector subsequently scores the feature vector using a fitted in-distribution distance, density, or robust z-score. The rationale aligns with the feature-extraction perspective of time-series analysis: rather than relying on a single specialised model, a broad array of interpretable temporal descriptors can summarise the structure of a time series \citep{fulcher2018}. This detector proves effective for gradual drift and temporal-pattern anomalies, although its efficacy is contingent upon the selected feature library and window length.

\subsection{Probabilistic ensemble dynamics model}

The probabilistic ensemble dynamics model (PEDM) detector fits an action-conditioned transition model,
\begin{equation}
 p_\theta(s_{t+1} \mid s_t, a_t),
\end{equation}
utilising an ensemble of probabilistic neural networks. At scoring time, each ensemble member samples candidate next-state predictions from its learned Gaussian transition distribution, and the anomaly score is the minimum sampled one-step prediction error,
\begin{equation}
 z_t =
 \min_{m \in \{1,\ldots,M\},\, k \in \{1,\ldots,K\}}
 \left\|s_{t+1} - \hat{s}^{(m,k)}_{t+1}\right\|_2^2,
 \qquad
 \hat{s}^{(m,k)}_{t+1} \sim p_{\theta_m}(\cdot \mid s_t,a_t),
\end{equation}
with the thresholded alarm calibrated as the maximum score observed on held-out in-distribution transitions. This conservative validation-maximum rule is not tuned directly for latency or recall. The detector draws direct inspiration from uncertainty-aware model-based RL \citep{chua2018} and RL-OOD work employing probabilistic dynamics models to identify severe MDP perturbations \citep{haider2023}. It is particularly pertinent when an anomaly alters the transition dynamics without rendering individual states obviously implausible.

\subsection{Autoencoder}

The autoencoder detector trains an encoder-decoder model on in-distribution inputs and scores reconstruction error,
\begin{equation}
 z_t = \|x_t - g_\phi(f_\theta(x_t))\|_2^2.
\end{equation}
The input $x_t$ may comprise a state, transition vector, or windowed trajectory feature. Reconstruction-based anomaly detection operates on the premise that a model trained exclusively on normal data reconstructs normal patterns more accurately than abnormal ones. This assumption is widely adopted in autoencoder-based anomaly detection and was evidenced in early neural anomaly-detection research by \citet{sakurada2014}. A limitation is that expressive autoencoders can sometimes reconstruct out-of-distribution inputs with low error; consequently, reconstruction error should not be treated as a calibrated probability of abnormality.

\subsection{Ensemble disagreement}

The ensemble-disagreement detector trains $M$ bootstrapped predictors and scores the variance of their outputs. For a transition predictor yielding outputs $\hat{s}^{(m)}_{t+1}$, a straightforward score is
\begin{equation}
 z_t = \frac{1}{M}\sum_{m=1}^{M}\left\|\hat{s}^{(m)}_{t+1} - \bar{s}_{t+1}\right\|_2^2,
 \qquad
 \bar{s}_{t+1} = \frac{1}{M}\sum_{m=1}^{M}\hat{s}^{(m)}_{t+1}.
\end{equation}
Deep ensembles constitute a simple and scalable uncertainty-estimation methodology \citep{lakshminarayanan2017}. In RL-OOD contexts, ensemble disagreement is valuable because models trained on identical in-distribution transitions often exhibit greater disagreement in regions with weak training support or shifted dynamics \citep{mohammed2021,haider2023}. Disagreement should be interpreted as an epistemic-uncertainty proxy rather than a comprehensive anomaly score.

\subsection{Cumulative sum}

The cumulative sum (CUSUM) detector monitors a scalar residual or feature statistic $e_t$, accumulating deviations from an in-distribution reference mean $\nu$. A one-sided formulation is
\begin{equation}
 g_t = \max(0, g_{t-1} + e_t - \nu - k),
 \qquad
 z_t = g_t,
\end{equation}
where $k$ represents a slack or drift parameter. The procedure follows the cumulative-sum change-detection method introduced by \citet{page1954} and subsequently formalised in broader abrupt-change detection treatments \citep{basseville1993}. Within OOD-RL-Bench, $e_t$ may constitute a state residual, reward residual, transition-model loss, or feature deviation. CUSUM is appropriate for sustained post-onset changes, yet it may prove less effective for brief point anomalies.

\subsection{KDE density}

The kernel density estimation (KDE) detector estimates an in-distribution density utilising a kernel estimator,
\begin{equation}
 \hat{p}(x) = \frac{1}{nh^d}\sum_{i=1}^{n}K\left(\frac{x-x_i}{h}\right),
\end{equation}
and scores low-density observations as anomalous via $z_t=-\log(\hat{p}(x_t)+\epsilon)$. Kernel density estimation is a classical non-parametric density-estimation technique \citep{parzen1962}. Its appeal lies in its avoidance of imposing a single Gaussian covariance model, rendering it complementary to Mahalanobis distance. However, dimensionality presents a limitation: KDE becomes computationally demanding in high-dimensional spaces unless features are carefully selected or compressed.

\subsection{Lyapunov divergence}

The Lyapunov-divergence detector estimates the local sensitivity of trajectories to proximate initial conditions. For embedded windows $y_t$, a local divergence score can be constructed from the growth of nearest-neighbour distances,
\begin{equation}
 z_t = \frac{1}{H}\sum_{h=1}^{H}\log\frac{\|y_{t+h}-y_{j+h}\|_2 + \epsilon}{\|y_t-y_j\|_2 + \epsilon},
\end{equation}
where $y_j$ represents a nearest neighbour of $y_t$ within an in-distribution reference set. Algorithms for estimating Lyapunov exponents from time series were developed to quantify divergence and chaos in reconstructed phase spaces \citep{wolf1985,kantz2004}. Within RL-OOD detection, this signal proves valuable when an anomaly alters local trajectory stability or phase-space geometry. The detector must be employed cautiously, as finite trajectories, stochasticity, and suboptimal embedding choices can produce unstable estimates.

\section{Anomaly-Injection Protocol}
\label{sec:injection}

OOD-RL-Bench employs two anomaly execution modes outlined in Table~\ref{tab:anomaly-execution-modes}. The detectors therefore continue to score trajectories offline; however, the data they score may originate from either a post-hoc transformation or a rollout subjected to a live fault.

\begin{table}[t]
\centering
\small
\caption{Execution modes used for anomaly injection.}
\label{tab:anomaly-execution-modes}
\begin{tabular}{@{}>{\raggedright\arraybackslash}p{0.08\linewidth}>{\raggedright\arraybackslash}p{0.30\linewidth}>{\raggedright\arraybackslash}p{0.20\linewidth}>{\raggedright\arraybackslash}p{0.30\linewidth}@{}}
\toprule
Mode & Rollout path & Concrete injectors & Rationale \\
\midrule
Post-hoc & A trained policy is executed in the clean environment to yield a stored trajectory, after which an injector modifies selected trajectory characteristics. & \texttt{gaussian state noise}; \texttt{state feature dropout}; \texttt{correlation break}; \texttt{regime switch}; \texttt{observation delay} & Faithful when the anomaly can be represented as a perturbation of the recorded observation stream. \\
\addlinespace[12pt]
Online & The policy is rolled out while the fault is applied during the environment interaction, and the resulting live trajectory is scored by the same fitted detectors. & \texttt{stuck actuator}; \texttt{dynamics scaling}; \texttt{action conditioned dynamics}; \texttt{gradual drift} & Required when the fault changes the policy--environment interaction and downstream state visitation. \\
\bottomrule
\end{tabular}
\end{table}

The recommended injector families and their concrete instantiations are summarised in Table~\ref{tab:injector-families}. These injectors are reference implementations rather than a closed perturbation set: new anomaly injectors can be incorporated by adhering to the appropriate post-hoc or online injector contracts and registering their parameters in the configuration. These families are selected to encompass sensor, temporal, action, transition, and distributional channels. Each injector exposes a scalar \emph{severity} parameter $\sigma \in [0,1]$ that controls perturbation magnitude in injector-specific units.

\begin{table}[t]
\centering
\small
\caption{Anomaly injector families and concrete instantiations.}
\label{tab:injector-families}
\begin{tabular}{@{}>{\raggedright\arraybackslash}p{0.25\linewidth}>{\raggedright\arraybackslash}p{0.39\linewidth}>{\raggedright\arraybackslash}p{0.26\linewidth}@{}}
\toprule
Injector family & Concrete injector(s) & Perturbation channel \\
\midrule
Observation noise & \texttt{gaussian state noise} & Sensor noise \\
Feature dropout & \texttt{state feature dropout} & Sensor dropout or telemetry loss \\
Observation delay & \texttt{observation delay} & Temporal latency \\
Actuator faults & \texttt{stuck actuator} & Action execution fault \\
Dynamics scaling & \texttt{dynamics scaling}; \texttt{action conditioned dynamics}; \texttt{gradual drift} & Transition dynamics; constant, action-conditioned, and ramped variants \\
Regime switch & \texttt{regime switch} & Distributional shift \\
Correlation breakdown & \texttt{correlation break} & Cross-coordinate and temporal coupling \\
\bottomrule
\end{tabular}
\end{table}

\subsection{Gaussian state noise}
The injector introduces independent Gaussian noise to each state coordinate at every post-onset step. The standard deviation for each coordinate is determined by $\sigma \cdot \mathrm{std}(s_{i,1:T})$, derived from the clean trajectory. Consequently, coordinates exhibiting low variance under the clean policy undergo proportionately smaller perturbations, thereby preserving their original signs and approximate magnitudes, which simulates the behaviour of inaccurate or noisy sensors.

\subsection{State feature dropout}
Each post-onset state coordinate is independently set to zero with probability $0.8\sigma$, which simulates intermittent sensor failure or telemetry loss. Coordinates that survive this process pass through unaltered. The presence of (near-)zero coordinate entries that the policy has rarely encountered at zero generates a strongly out-of-distribution joint state.

\subsection{Correlation break}
The injector selects $\lceil \sigma \cdot d_s \rceil$ state dimensions and shuffles their values across time within the post-onset window. While the marginal distribution for each dimension is preserved, the joint coupling between coordinates is disrupted. This mimics sensor cross-talk or a software defect that causes confusion in which reading corresponds to which signal. Although each selected coordinate retains a plausible distribution, the temporal and cross-coordinate relationships no longer align with the dynamics that generated the clean trajectory

\subsection{Stuck actuator}
After $t^\star$, this online injector intermittently fixes the executed action to a held post-onset action, with probability controlled by severity $\sigma$. The policy continues to emit commanded actions, which are logged within the trajectory's action channel; the executed action is logged separately in the trajectory metadata. At $\sigma=1$, the actuator remains fully stuck until the episode resets. Within a discrete action space this emulates an actuator that increasingly ignores the policy command and repeats a held action. The ensuing state trajectory diverges from the original clean rollout through the closed-loop effects of repeatedly applying an incorrect action. This explicitly justifies online evaluation of the anomaly, rather than retrospective alteration of action labels.

\subsection{Dynamics scaling}
During a live rollout, post-onset state increments $\Delta s_t = s_{t+1} - s_t$ are multiplied by a constant scale factor of $1 + 1.5\sigma$ and written back into the environment state before the next policy decision. This simulates an abrupt change in effective mass, friction, or actuator gain (e.g., a terrain alteration or shift in payload), while allowing the policy's subsequent actions to depend on the altered state. The resulting trajectory drifts onto a manifold upon which the policy was never trained.

\subsection{Action conditioned dynamics}
This online injector is identical to dynamics scaling except that the increment scaling is applied only on timesteps where the commanded action equals a fixed target action $a^\star$ chosen by the scenario configuration. The live trajectory is unmodified on all other timesteps. This injector serves as the most rigorous test of action-conditioned dynamics models: marginal-state detectors observe a trajectory where the displacement increments are intermittently anomalous, whereas a detector that conditions on the action receives a coherent signal precisely every time $a_t = a^\star$.

\subsection{Gradual drift}
During a live rollout, the dynamics scale increases linearly from $1.0$ at $t^\star$ to $1 + 1.5\sigma$ across $\mathrm{ramp\_steps} = 200$ steps, remaining constant thereafter. This models progressive mechanical degradation, such as lubricant depletion or joint stiffening. The initial post-onset period remains nearly indistinguishable from clean dynamics, meaning detectors reliant on instantaneous outliers register no anomaly. The trajectory only diverges from the in-distribution support as the ramp progresses, providing a structurally advantageous signal for change-detection algorithms such as CUSUM.

\subsection{Regime switch}
At $t^\star$, the post-onset state distribution is displaced by $2.0\sigma$ pre-onset standard deviations, while the per-coordinate variance is amplified by a factor of $1 + 2.0\sigma$. Contrary to dynamics-based injectors, the state sequence is not re-propagated under the policy; instead, the marginal observation distribution abruptly shifts to a novel operating regime. Density- and Mahalanobis-style detectors capture this near-perfectly, as the post-onset marginal is displaced outside the fitted in-distribution support.

\subsection{Observation delay}
Post-onset observations are substituted post hoc with the state from $k$ steps prior, where $k = \lceil \sigma \cdot \mathrm{max\_delay} \rceil$ and $\mathrm{max\_delay} = 10$. This simulates a communication or sensor-processing latency in the recorded observation stream without altering the policy's original clean rollout. Because the delayed states remain on the in-distribution manifold, marginal- and density-based detectors remain blind; only detectors that juxtapose the observed state against temporal or action-conditioned context receive a coherent signal.

\section{Evaluation Methodology}
\label{sec:evaluation}

\subsection{Evaluation Metrics}
\label{sec:metrics}
Each detector yields a score sequence $z_{0:T-1}$. The principal threshold-free metric is AUROC, which assesses ranking quality across all potential thresholds and is extensively utilised for binary discrimination tasks \citep{fawcett2006}. Given that anomaly labels are frequently imbalanced, AUPRC should also be reported where performance on the positive class is paramount \citep{davis2006}. Thresholded metrics must encompass F1, precision, recall, and the false-positive rate at a fixed true-positive-rate target. Furthermore, temporal metrics should comprise detection delay subsequent to the known anomaly onset and segment-level overlap measures.

The benchmark must also report a null baseline. In the null run, clean trajectories from the same execution mode are scored against a counterfactual label vector that designates a post-onset interval as anomalous, despite the absence of an injected anomaly. If a detector's score naturally increases towards the end of an episode, the null AUROC will exceed random chance. The adjusted score
\begin{equation}
 \widehat{\auroc}_{\mathrm{adj}} = \min\left(1, \max\left(0, \auroc_{\mathrm{raw}} - \auroc_{\mathrm{null}} + 0.5\right)\right)
\end{equation}
should be reported in conjunction with the raw AUROC. This correction does not act as a standard substitute for AUROC but as a benchmark-specific diagnostic for score drift under clean trajectories. 

The empirical results below use \emph{matched-time negatives} as the primary threshold-free comparison. For each perturbed episode, the positive class contains raw anomalous scores at post-onset timesteps, $\{z^{\mathrm{ood}}_t : y_t=1\}$. The negative class contains raw clean scores from the same execution mode, seed, and timestep indices, $\{z^{\mathrm{clean}}_t : y_t=1\}$. AUROC, AUPRC, and $\fpr_{0.95}$ are then computed on this balanced clean-versus-anomalous score set. This protocol does not define an episode-level alarm time, so detection delay remains the delay of the detector's calibrated threshold on the original perturbed trajectory.

\subsection{Experimental Design}
\label{sec:experimental_design}

A valid experiment requires all configured detectors to be fit on the same set of in-distribution trajectories and to be evaluated against the same execution protocol for each injector. The basic unit of reproducible reporting is a single row for each detector--anomaly pair, containing the execution mode, raw AUROC, null-adjusted AUROC, matched-time AUROC, matched-time AUPRC, matched-time \(\fpr_{0.95}\), F1, detection delay, segment precision, and segment recall. The experiment must also document the number of in-distribution trajectories, the number of out-of-distribution trajectories, episode length, anomaly onset, detector hyperparameters, random seeds, environment version, policy checkpoint, software versions, whether each anomaly was generated post hoc or online, and the confidence-interval method used for each reported metric.

\subsection{Expected Detector Behaviour}
\label{sec:expected_behaviour}

The detector suite is intentionally heterogeneous. \texttt{mahalanobis distance} and \texttt{kde density} are expected to perform best when anomalies displace observations or transition features from the fitted in-distribution support. \texttt{pedm} and \texttt{ensemble disagreement} are designed to respond when anomalies alter action-conditioned transition dynamics. \texttt{autoencoder} should remain competitive when abnormal inputs are characterised by poor reconstruction fidelity, though it may fail if the learned representation generalises too extensively. \texttt{cusum} should perform best for sustained shifts in a selected residual statistic. \texttt{time series features} should identify anomalies that perturb local temporal structure but may be sensitive to feature selection. \texttt{lyapunov divergence} is anticipated to be useful for shifts in local stability or phase-space geometry but requires careful interpretation, given that Lyapunov estimates are sensitive to noise, finite data, and clean temporal score trends \citep{wolf1985,kantz2004}.

\subsection{Evaluation Protocol}
\label{sec:protocol}

The assessment framework proceeds in five distinct phases:
\begin{enumerate}
    \item A static policy is executed within an unperturbed environment to generate $N_{\mathrm{ID}}$ in-distribution trajectories, each of duration $T$, which serve as the calibration set for each detector $D$.
    \item $N_{\mathrm{OOD}}$ supplementary trajectories are gathered for every injector $I_j$: post-hoc injectors first collect a clean trajectory and then perturb its recorded fields, whereas online injectors perturb the live environment during rollout and separately collect a clean trajectory with the same seed for matched-time comparison. For online faults, the matched clean rollout is a same-seed no-fault counterfactual matched by episode time, not by identical post-onset state. Because the fault changes the closed-loop policy--environment interaction, the perturbed and clean trajectories may diverge after $t^\star$; matched-time metrics therefore control for temporal position but intentionally include the downstream trajectory consequences of the live fault.
    \item This process yields a perturbed trajectory $\tilde{\tau}$, a matched clean trajectory $\tau^{\mathrm{clean}}$, and a binary ground-truth annotation vector $y_t = \mathbbm{1}\{t \ge t^\star\}$. Each calibrated detector then assigns anomaly scores to both trajectories.
    \item Performance is quantified in the benchmark output using threshold-free matched-time metrics (AUROC, AUPRC), matched-time $\fpr_{0.95}$, threshold-dependent trajectory metrics (F1 and detection delay), and temporal/segment metrics (temporal AUROC, segment precision, and segment recall). The analysis below focuses on matched-time AUROC/AUPRC, matched-time $\fpr_{0.95}$, detection delay, and null-baseline diagnostics. These statistics are calculated individually per episode and aggregated as mean~$\pm$~standard deviation across all $N_{\mathrm{OOD}}$ episodes.
    \item Baseline evaluations replicate the scoring process on clean post-hoc and clean online trajectories, utilising both an all-zero clean label vector for thresholded false-positive rates and a counterfactual annotation vector $y^{\mathrm{null}}_t = \mathbbm{1}[t \ge t^\star]$ for temporal-bias diagnostics. Matched-time AUROC, AUPRC, and $\fpr_{0.95}$ confidence intervals use episode bootstrap over paired clean/anomalous episodes; lift-over-null and null-adjusted diagnostic metrics use paired episode bootstrap over matched null/OOD seeds.
\end{enumerate}

Detection delay is calculated as the temporal offset between $t^\star$ and the initial timestamp exceeding the classification threshold following $t^\star$; episodes where the detector fails to trigger are recorded at the terminal timestep of the trajectory. At the segment level, each continuous run of thresholded positives is consolidated into a solitary detection instance, with precision and recall assessed relative to the ground-truth interval $[t^\star, T-1]$. The anomaly-score threshold used to convert continuous scores into binary anomaly labels is calibrated on in-distribution data according to each detector's rule: most detectors use a high clean-score percentile, whereas PEDM uses the maximum held-out validation score.

\section{Demonstration Experiment}
\label{sec:setup}

We report here an experiment designed to demonstrate the utility of the OOD-RL-Bench framework. The aim is not to carry out a comprehensive investigation of OOD detectors, but instead to outline the ability of OOD-RL-Bench to highlight the impact which different types of anomaly may have on the performance of detectors.

\subsection{Environment and policy}
The benchmark is applied to LunarLander-v3, used here through Gymnasium \citep{brockman2016,towers2024}, utilising its eight-dimensional flat-state observation space and four discrete actions. The state vector is $(x, y, v_x, v_y, \theta, \omega, c_1, c_2)$, where $(x,y)$ is the lander position, $(v_x, v_y)$ its linear velocity, $\theta$ its tilt, $\omega$ its angular velocity, and $c_1, c_2 \in \{0,1\}$ are the two leg-contact flags. Actions are discrete: $\{0$: no-op, $1$: left engine, $2$: main engine, $3$: right engine$\}$. The control policy employs a NoisyNet DQN architecture \citep{mnih2015} equipped with an MLP value head, optimised using the flat state representation over 210{,}000 environment steps. Hyperparameters include a batch size of 64, a discount factor of 0.99, a learning rate of $5\times 10^{-4}$, a replay buffer capacity of $10^{5}$, and target-network synchronisation every 1{,}000 steps. Training was concluded by satisfying an early stopping rule; the final checkpoint achieves a mean return of approximately 230 over 30 episodes, consistent with a fully solved LunarLander policy. This identical checkpoint governs all in-distribution and out-of-distribution rollouts; the policy undergoes no further retraining or fine-tuning on perturbed data.

\subsection{Benchmark configuration}
Each rollout is limited to a maximum of $T = 500$ steps. Detector fitting relies on thirty in-distribution trajectories ($N_{\mathrm{ID}} = 30$), while evaluation utilises twenty perturbed trajectories per anomaly injector ($N_{\mathrm{OOD}} = 20$). The anomaly onset is fixed at $t^\star = \lfloor T/3 \rfloor = 166$, ensuring that every episode includes a substantial pre-onset phase and an extended post-onset phase. Nine anomaly injectors are assessed. Five are evaluated post hoc: \texttt{gaussian state noise}, \texttt{state feature dropout}, \texttt{correlation break}, \texttt{regime switch}, and \texttt{observation delay}. Four are evaluated online in the live environment: \texttt{stuck actuator}, \texttt{dynamics scaling}, \texttt{action conditioned dynamics}, and \texttt{gradual drift}. In the LunarLander configuration, observation-noise perturbations affect all eight state components; feature dropout is capable of zeroing any coordinate; correlation break permutes severity-selected coordinates across time; and observation delay substitutes earlier state vectors while preserving the recorded action stream. The online faults operate via the four-action engine interface: a stuck actuator repeats a held engine action, dynamics scaling writes scaled increments back into the simulator state, action-conditioned dynamics targets the main-engine action ($a^\star = 2$), and gradual drift ramps the same scale factor across 200 steps. Injector-specific severity parameters are documented within the configuration file distributed with the framework. 

\subsection{Reproducibility}
An integer seed sets the environment, policy noise, in-distribution data collection, and detector initialisation across all runs. Out-of-distribution rollouts employ this base seed augmented by a fixed offset, thereby ensuring independence from the calibration set while preserving determinism. Hardware and software specifications are documented for each run: the results were produced with Python 3.13.9, PyTorch 2.9.1, NumPy 2.4.0, and MPS execution. The framework source code, the trained policy checkpoint, configuration files, and the comprehensive per-condition results files are published alongside this manuscript.

\section{Results}
\label{sec:results}

\subsection{Per-detector averages}
\label{sec:per-detector}

Table \ref{tab:per-detector} displays the mean matched-time AUROC and AUPRC for each configured detector, calculated as a uniform average across the nine anomaly scores in the mixed post-hoc/online protocol. \texttt{pedm} takes the lead with the highest aggregate matched-time AUROC (0.874), followed by \texttt{time series features} (0.839), \texttt{kde density} (0.730), and \texttt{ensemble disagreement} (0.720). This ranking uses each detector's raw score scale; the only control is that anomalous post-onset scores are compared against clean scores from the same timesteps. \texttt{lyapunov divergence}, which achieves a high raw aggregate AUROC by separating clean scores between early and late timesteps, falls to near-chance matched-time performance (0.539). The AUPRC is directly interpretable against a balanced no-skill baseline of approximately 0.5, as each post-onset anomaly score is paired with a corresponding clean score from the same timestep.

\begin{table}[t]
\centering
\small
\caption{Mean matched-time AUROC and AUPRC per detector, averaged uniformly across the nine anomaly rows (LunarLander-v3, 20 OOD episodes per injector). Detectors sorted by AUROC.}
\label{tab:per-detector}
\begin{tabular}{lcc}
\toprule
Detector & Mean AUROC & Mean AUPRC \\
\midrule
\texttt{pedm} & 0.874 & 0.851 \\
\texttt{time series features} & 0.839 & 0.847 \\
\texttt{kde density} & 0.730 & 0.743 \\
\texttt{ensemble disagreement} & 0.720 & 0.740 \\
\texttt{autoencoder} & 0.705 & 0.717 \\
\texttt{mahalanobis distance} & 0.649 & 0.675 \\
\texttt{lyapunov divergence} & 0.539 & 0.530 \\
\texttt{cusum} & 0.495 & 0.556 \\
\bottomrule
\end{tabular}
\end{table}

\subsection{Per-anomaly difficulty}
Anomaly difficulty is summarised by the matched-time AUROC of the top-performing detector for each injector (Table~\ref{tab:per-anomaly}). Three post-hoc observation injectors remain nearly saturated: \texttt{gaussian state noise}, \texttt{state feature dropout}, and \texttt{correlation break} all reach AUROC values at or above 0.998. \texttt{regime switch}, \texttt{stuck actuator}, \texttt{dynamics scaling}, and \texttt{gradual drift} also become high-accuracy rows. In the current eight-detector suite, \texttt{observation delay} remains moderate, and \texttt{action conditioned dynamics} remains the hardest action/dynamics row.

\begin{table}[t]
\centering
\small
\caption{Best detector (by mean matched-time AUROC) and the corresponding mean matched-time AUROC and AUPRC per anomaly injector under the mixed protocol, averaged across 20 OOD episodes.}
\label{tab:per-anomaly}
\begin{tabular}{lllcc}
\toprule
Anomaly & Mode & Best detector & AUROC & AUPRC \\
\midrule
\texttt{gaussian state noise} & post-hoc & \texttt{time series features} & 0.999 & 0.999 \\
\texttt{state feature dropout} & post-hoc & \texttt{time series features} & 0.998 & 0.998 \\
\texttt{correlation break} & post-hoc & \texttt{time series features} & 0.999 & 0.999 \\
\texttt{regime switch} & post-hoc & \texttt{kde density} & 0.997 & 0.996 \\
\texttt{observation delay} & post-hoc & \texttt{pedm} & 0.697 & 0.692 \\
\texttt{stuck actuator} & online & \texttt{kde density} & 0.919 & 0.941 \\
\texttt{dynamics scaling} & online & \texttt{pedm} & 0.927 & 0.883 \\
\texttt{action conditioned dynamics} & online & \texttt{pedm} & 0.636 & 0.597 \\
\texttt{gradual drift} & online & \texttt{pedm} & 0.869 & 0.836 \\
\bottomrule
\end{tabular}
\end{table}

\subsection{Best detector per anomaly}
\label{sec:best-per}
No single detector monopolises the entire suite. \texttt{time series features} excels across the three simplest post-hoc observation perturbations, \texttt{kde density} tops \texttt{regime switch} and \texttt{stuck actuator}, and \texttt{pedm} leads \texttt{observation delay}, \texttt{dynamics scaling}, \texttt{action conditioned dynamics}, and \texttt{gradual drift}. This distribution aligns with the taxonomy detailed in Section~\ref{sec:detectors}: temporal-feature methods prove effective when the observable trajectory structure alters, density methods respond to marginal state-distribution shifts, and action-conditioned transition models assist when the perturbation modifies the transition context. The weak aggregate matched-time score for \texttt{lyapunov divergence} (mean AUROC 0.539, mean AUPRC 0.530) confirms that its strong raw AUROC is primarily a temporal-position artefact rather than evidence of effective same-time anomaly separation. Consequently, selecting a solitary detector for practical deployment necessitates a deliberate compromise: securing robust detection across one anomaly family inherently results in insensitivity to anomalies of a different type.

\subsection{Expected versus observed detector behaviour}
\label{sec:expected-observed}

The empirical results broadly corroborate the expectations detailed in Section~\ref{sec:expected_behaviour}, while simultaneously highlighting where they are incomplete. As predicted, \texttt{kde density} demonstrates superior performance when a fault induces a marginal shift in the state distribution, outperforming both \texttt{regime switch} and the resultant downstream state shift caused by \texttt{stuck actuator}. Similarly, \texttt{pedm} proves most effective with the online transition faults of \texttt{dynamics scaling}, \texttt{action conditioned dynamics}, and \texttt{gradual drift}; its dominance in \texttt{observation delay} aligns with an action-conditioned transition perspective, given that the delayed observation becomes incoherent with the concurrent action and next state context. The \texttt{time series features} approach exceeds the expectation of detecting only temporal-pattern anomalies; by capturing rolling variations in noise, dropout, and cross-coordinate structure, it leads across all three post-hoc observation perturbations. Conversely, \texttt{mahalanobis distance} fails to dominate these, despite its expected sensitivity to displaced features, which suggests that covariance-normalised distance is less effective than windowed temporal descriptors for the non-Gaussian and correlation-breaking perturbations employed in this study.

The primary deviations from the anticipated outcomes are related to calibration and temporal confounding, rather than the broad detector taxonomy. While \texttt{ensemble disagreement} yields low detection delays, it does not match the matched-time ranking of \texttt{pedm} across transition faults; \texttt{cusum} fails to emerge as the best gradual-drift detector under matched-time AUROC, despite being conceptually suited to sustained changes; \texttt{autoencoder} remains competitive but does not lead any anomaly category; and \texttt{lyapunov divergence} serves chiefly as a cautionary example, given that its raw separation is significantly influenced by clean temporal score trends. These results suggest that while expected detector behaviour provides a valuable prior, the benchmark remains essential for exposing how representation selection, score calibration, and matched-time controls influence the realised rankings.

\subsection{Severity-sweep capability}
\label{sec:severity-sweep}
Beyond the aggregate anomalies, the framework facilitates the evaluation of a fixed injector across a grid of severity levels. Table~\ref{tab:severity-sweep-example} provides one such sweep for \texttt{dynamics scaling} using \texttt{pedm}. With the detector and anomaly family held constant, the severity effect is visible: matched-time AUROC and AUPRC increase as the injected dynamics perturbation intensifies, while $\fpr_{0.95}$ decreases. This analysis complements the preceding detector-ranking tables by examining whether a benchmark condition becomes progressively more manageable as the magnitude of the perturbation increases.

\begin{table}[t]
\centering
\small
\caption{Example severity sweep for \texttt{dynamics scaling} with \texttt{pedm}. Values are mean matched-time AUROC, AUPRC, and $\fpr_{0.95}$ over 20 OOD episodes at each severity. Higher AUROC/AUPRC and lower $\fpr_{0.95}$ indicate easier detection.}
\label{tab:severity-sweep-example}
\begin{tabular}{cccc}
\toprule
Severity & AUROC & AUPRC & $\fpr_{0.95}$ \\
\midrule
0.10 & 0.608 & 0.570 & 0.695 \\
0.20 & 0.714 & 0.644 & 0.564 \\
0.40 & 0.826 & 0.755 & 0.397 \\
0.60 & 0.882 & 0.822 & 0.287 \\
0.80 & 0.927 & 0.883 & 0.170 \\
\bottomrule
\end{tabular}
\end{table}

\subsection{Detection delay}
\label{sec:delay}
Threshold-free matched-time metrics assess the ability of a detector to \emph{rank} anomalous post-onset scores against same-time clean scores, yet they offer no information regarding the timing of an episode-level alarm. Detection delay serves as the complementary temporal metric: it measures the number of steps elapsed between the ground-truth onset $t^\star$ and the first thresholded positive emitted by the detector. Episodes where the detector fails to fire are recorded at the trajectory horizon ($T - t^\star = 334$ steps); consequently, a detector that achieves a strong AUROC score but rarely crosses its calibrated threshold will exhibit a substantial delay, which aligns with the intended behaviour of this metric.

Table~\ref{tab:delay-per-detector} outlines the mean detection delay for the calibrated raw detector thresholds across the nine rows. This metric remains distinct from matched-time AUROC because the matched-time comparison acts as a pairwise discrimination test rather than a deployment alarm. \texttt{time series features} achieves the lowest average latency (49.2 steps), followed by \texttt{mahalanobis distance} (108.9 steps), \texttt{autoencoder} (130.3 steps), and \texttt{ensemble disagreement} (131.6 steps). \texttt{pedm} maintains the highest matched-time AUROC but suffers from a slow raw-threshold delay (271.9 steps). For \texttt{pedm}, this delay arises because the alarm threshold is set to the maximum held-out in-distribution validation score: while many anomalous scores exceed their same-time clean counterparts, they often remain below this conservative threshold until late in the episode. This highlights that ranking quality and calibrated alarm timing are separate characteristics.

\begin{table}[t]
\centering
\small
\caption{Mean raw-threshold detection delay per detector (in steps), aggregated across the nine anomaly rows. The trajectory horizon is $T - t^\star = 334$ steps; values close to this ceiling indicate that the detector frequently never fires within the episode.}
\label{tab:delay-per-detector}
\begin{tabular}{lccc}
\toprule
Detector & mean & min & max \\
\midrule
\texttt{time series features} & 49.2 & 0.0 & 224.7 \\
\texttt{mahalanobis distance} & 108.9 & 0.0 & 266.6 \\
\texttt{autoencoder} & 130.3 & 0.2 & 282.2 \\
\texttt{ensemble disagreement} & 131.6 & 0.0 & 293.5 \\
\texttt{cusum} & 145.5 & 5.6 & 314.3 \\
\texttt{kde density} & 163.1 & 0.0 & 303.9 \\
\texttt{lyapunov divergence} & 261.6 & 153.9 & 334.0 \\
\texttt{pedm} & 271.9 & 115.0 & 334.0 \\
\bottomrule
\end{tabular}
\end{table}

Table~\ref{tab:delay-per-anomaly} summarises the raw-threshold delay associated with the detector achieving the highest matched-time AUROC for each injector. The ``fastest'' column indicates the minimum delay attainable across the eight-detector suite, alongside the matched-time AUROC score corresponding to that latency. For the simpler post-hoc observation injectors and \texttt{regime switch}, the detector with the best ranking is either the fastest available or within a single step of the fastest detector. Conversely, for the online dynamics rows, the detectors that rank best under matched-time AUROC may still be slow to trigger under the raw calibrated threshold; the faster detectors in these scenarios often sacrifice matched-time AUROC performance to achieve that latency.

\begin{table}[t]
\centering
\small
\caption{Mean raw-threshold detection delay (steps after $t^\star$), averaged across 20 OOD episodes, under two detector-selection rules per anomaly injector. ``Best AUROC'' reports the mean delay and mean matched-time AUROC of the detector with the highest mean matched-time AUROC; ``Fastest'' reports the detector with the lowest mean delay together with the mean matched-time AUROC it pays for that latency. Trajectory horizon $= 334$ steps.}
\label{tab:delay-per-anomaly}
\begin{tabular}{lcc cc}
\toprule
 & \multicolumn{2}{c}{Best AUROC} & \multicolumn{2}{c}{Fastest} \\
\cmidrule(lr){2-3} \cmidrule(lr){4-5}
Anomaly & detector & delay (AUROC) & detector & delay (AUROC) \\
\midrule
\texttt{gaussian state noise} & \texttt{tsf} & 0.1 (0.999) & \texttt{tsf} & 0.1 (0.999) \\
\texttt{state feature dropout} & \texttt{tsf} & 0.9 (0.998) & \texttt{tsf} & 0.9 (0.998) \\
\texttt{correlation break} & \texttt{tsf} & 0.0 (0.999) & \texttt{tsf} & 0.0 (0.999) \\
\texttt{regime switch} & \texttt{kde} & 0.0 (0.997) & \texttt{mah} & 0.0 (0.981) \\
\texttt{observation delay} & \texttt{pedm} & 322.8 (0.697) & \texttt{tsf} & 72.6 (0.533) \\
\texttt{stuck actuator} & \texttt{kde} & 45.3 (0.919) & \texttt{mah} & 32.1 (0.889) \\
\texttt{dynamics scaling} & \texttt{pedm} & 297.3 (0.927) & \texttt{tsf} & 26.9 (0.899) \\
\texttt{action conditioned dynamics} & \texttt{pedm} & 334.0 (0.636) & \texttt{cusum} & 145.9 (0.470) \\
\texttt{gradual drift} & \texttt{pedm} & 326.1 (0.869) & \texttt{tsf} & 81.3 (0.799) \\
\bottomrule
\end{tabular}
\\[2pt]
\footnotesize \texttt{tsf} = \texttt{time series features}, \texttt{kde} = \texttt{kde density}, \texttt{mah} = \texttt{mahalanobis distance}.
\end{table}

A consistent trend visible across both tables indicates that low latency does not equate to reliable ranking. \texttt{time series features} and \texttt{cusum} frequently fire ahead of the detector with the best ranking, yet their matched-time AUROC performance is weaker across several online dynamics rows. Conversely, while \texttt{pedm} demonstrates strong ranking capabilities for specific online dynamics anomalies under matched-time AUROC, its raw threshold (derived from the validation maximum) can still result in delayed activation. The most prominent exception is \texttt{dynamics scaling}, where \texttt{time series features} achieves an AUROC nearly comparable to \texttt{pedm} while activating considerably sooner. Consequently, a practical approach to detector selection necessitates consideration of both ranking metrics and calibrated alarm analysis.

\subsection{False-positive rate at 95\% TPR}
\label{sec:fpr95}
Although AUROC captures ranking quality across the full range of potential thresholds, a deployed detector functions at a single threshold and suffers a false-alarm penalty when clean scores exceed it. The false-positive rate evaluated at $95\%$ true-positive rate ($\fpr_{0.95}$) is a commonly reported operating-point metric in OOD detection benchmarks \citep{liang2018,yang2022,zhang2023}. This quantifies the proportion of same-time clean scores that trigger an alarm when the threshold is calibrated to capture $95\%$ of anomalous post-onset scores. Operationally, the threshold is determined from the matched-time anomalous score distribution as the largest threshold maintaining at least $95\%$ TPR; $\fpr_{0.95}$ is then the fraction of matched clean scores exceeding that same threshold. Lower values are desirable; perfect separation yields $0.000$, whereas an uninformative detector results in approximately $0.95$.  

Table~\ref{tab:fpr-per-detector} summarises the matched-time $\fpr_{0.95}$ for each detector across the nine anomaly rows. \texttt{pedm} achieves the lowest aggregate $\fpr_{0.95}$ (0.378), closely followed by \texttt{time series features} (0.384). A value near 0.95 indicates an uninformative operating point under the balanced matched-time comparison. \texttt{cusum}, \texttt{mahalanobis distance}, and \texttt{lyapunov divergence} exhibit weak aggregate operating points, even when their AUROC remains useful for specific anomaly families.

\begin{table}[t]
\centering
\small
\caption{Mean matched-time false-positive rate at $95\%$ TPR per detector, aggregated across the nine anomalies. Lower is better; an uninformative detector reads approximately $0.95$.}
\label{tab:fpr-per-detector}
\begin{tabular}{lccc}
\toprule
Detector & mean & min & max \\
\midrule
\texttt{pedm} & 0.378 & 0.008 & 0.876 \\
\texttt{time series features} & 0.384 & 0.000 & 0.945 \\
\texttt{kde density} & 0.609 & 0.006 & 0.947 \\
\texttt{ensemble disagreement} & 0.642 & 0.062 & 0.944 \\
\texttt{autoencoder} & 0.668 & 0.033 & 0.949 \\
\texttt{lyapunov divergence} & 0.742 & 0.444 & 0.957 \\
\texttt{mahalanobis distance} & 0.743 & 0.116 & 0.952 \\
\texttt{cusum} & 0.806 & 0.127 & 0.992 \\
\bottomrule
\end{tabular}
\end{table}

Table~\ref{tab:fpr-per-anomaly} details matched-time $\fpr_{0.95}$ for the detector that achieves the highest matched-time AUROC on each injector, as well as the detector that independently minimises $\fpr_{0.95}$. The selections align for eight of the nine injectors; the exception being \texttt{action conditioned dynamics}, where \texttt{time series features} has a lower false-positive rate than \texttt{pedm} but also lower AUROC. The hard channels retain high operating-point error rates even after choosing the best detector.

\begin{table}[t]
\centering
\small
\caption{Mean matched-time $\fpr_{0.95}$, averaged across 20 OOD episodes, under two detector-selection rules per anomaly injector. ``Best AUROC'' reports the mean $\fpr_{0.95}$ and mean matched-time AUROC of the detector with the highest mean matched-time AUROC; ``Best $\fpr_{0.95}$'' reports the detector that minimises mean $\fpr_{0.95}$ together with the mean matched-time AUROC it offers.}
\label{tab:fpr-per-anomaly}
\begin{tabular}{lcc cc}
\toprule
 & \multicolumn{2}{c}{Best AUROC} & \multicolumn{2}{c}{Best $\fpr_{0.95}$} \\
\cmidrule(lr){2-3} \cmidrule(lr){4-5}
Anomaly & detector & $\fpr_{0.95}$ (AUROC) & detector & $\fpr_{0.95}$ (AUROC) \\
\midrule
\texttt{gaussian state noise} & \texttt{tsf} & 0.000 (0.999) & \texttt{tsf} & 0.000 (0.999) \\
\texttt{state feature dropout} & \texttt{tsf} & 0.000 (0.998) & \texttt{tsf} & 0.000 (0.998) \\
\texttt{correlation break} & \texttt{tsf} & 0.000 (0.999) & \texttt{tsf} & 0.000 (0.999) \\
\texttt{regime switch} & \texttt{kde} & 0.006 (0.997) & \texttt{kde} & 0.006 (0.997) \\
\texttt{observation delay} & \texttt{pedm} & 0.846 (0.697) & \texttt{pedm} & 0.846 (0.697) \\
\texttt{stuck actuator} & \texttt{kde} & 0.509 (0.919) & \texttt{kde} & 0.509 (0.919) \\
\texttt{dynamics scaling} & \texttt{pedm} & 0.170 (0.927) & \texttt{pedm} & 0.170 (0.927) \\
\texttt{action conditioned dynamics} & \texttt{pedm} & 0.876 (0.636) & \texttt{tsf} & 0.764 (0.489) \\
\texttt{gradual drift} & \texttt{pedm} & 0.487 (0.869) & \texttt{pedm} & 0.487 (0.869) \\
\bottomrule
\end{tabular}
\\[2pt]
\footnotesize \texttt{tsf} = \texttt{time series features}, \texttt{kde} = \texttt{kde density}.
\end{table}

When evaluated alongside the AUROC and detection-delay outcomes, $\fpr_{0.95}$ provides the final component of the analysis. In the case of simpler injectors, all three metrics converge on a limited selection of dominant detectors. Conversely, for live dynamics anomalies and complex post-hoc scenarios, the matched-time AUROC, $\fpr_{0.95}$, and delay continue to diverge: a detector may effectively distinguish same-time clean and anomalous scores yet struggle to surpass its calibrated deployment threshold in a timely manner. Consequently, selecting a deployment detector for these contexts necessitates a deliberate commitment to a definitive operational priority—whether ranking performance, a calibrated false-alarm threshold, or latency—rather than relying solely on AUROC.

\subsection{Null baseline}
The null baseline is derived independently for both post-hoc and online execution modes by evaluating clean trajectories against all-zero clean labels alongside the identical counterfactual post-onset label vector utilised for temporal-bias diagnostics. With the LunarLander trial, these clean post-hoc and clean online baselines coincide, as the no-fault live collector reproduces the same clean policy and environment interaction. Table~\ref{tab:null-baseline-auroc} reports the corresponding raw counterfactual null AUROC values for the eight-detector suite, using the same LunarLander sweep output as the headline results. These threshold-free null metrics underscore the rationale for preferring matched-time negatives in rankings: \texttt{lyapunov divergence} attains a raw counterfactual null AUROC of 0.812, while \texttt{pedm}, \texttt{ensemble disagreement}, \texttt{mahalanobis distance}, \texttt{time series features}, and \texttt{cusum} demonstrate raw null AUROC values between 0.64 and 0.68. As the matched-time comparison aligns post-onset anomaly scores with the corresponding post-onset clean scores, it controls the temporal-position effect via the use of raw same-time clean negatives. The matched-time means yield 0.539 for \texttt{lyapunov divergence} and 0.874 for \texttt{pedm}, contrasting sharply with their raw aggregate AUROC figures of 0.901 and 0.935 respectively.

\begin{table}[t]
\centering
\small
\caption{Raw counterfactual null AUROC on clean LunarLander trajectories, computed by applying the post-onset label vector to unperturbed rollouts. In this deterministic LunarLander run, the clean post-hoc and no-fault online control rollouts coincide, so a single null-baseline value is reported per detector.}
\label{tab:null-baseline-auroc}
\begin{tabular}{lccc}
\toprule
Detector & AUROC & SD & 95\% CI \\
\midrule
\texttt{lyapunov divergence} & 0.812 & 0.122 & [0.761, 0.865] \\
\texttt{pedm} & 0.675 & 0.117 & [0.624, 0.726] \\
\texttt{ensemble disagreement} & 0.669 & 0.120 & [0.623, 0.712] \\
\texttt{mahalanobis distance} & 0.661 & 0.219 & [0.578, 0.748] \\
\texttt{time series features} & 0.642 & 0.183 & [0.577, 0.713] \\
\texttt{cusum} & 0.637 & 0.257 & [0.528, 0.738] \\
\texttt{autoencoder} & 0.457 & 0.174 & [0.391, 0.530] \\
\texttt{kde density} & 0.376 & 0.253 & [0.267, 0.492] \\
\bottomrule
\end{tabular}
\end{table}

\subsection{Interpreting AUROC versus AUPRC}
\label{sec:auroc-vs-auprc}
The two threshold-free metrics measure distinct properties of the score sequence. AUROC integrates the true-positive rate against the false-positive rate and remains invariant to class prevalence: a wholly uninformative detector obtains a score of $0.5$, as the metric relies exclusively on the relative ranking of positive and negative scores. Conversely, AUPRC integrates precision against recall and is inherently sensitive to prevalence. Under the matched-time headline protocol, the positive and negative classes are intrinsically balanced: every post-onset anomalous score corresponds to one clean score from the same timestep. Consequently, a non-informative scoring strategy yields an AUPRC close to $0.5$, rather than the post-onset episode-label prior of approximately $0.668$ applicable to raw pre-versus-post trajectory metrics.

This characteristic renders the matched-time AUPRC table more intuitive to interpret than the raw trajectory table. Both metrics approach perfect separation for the simplest observation perturbations: \texttt{time series features} achieves matched-time $\auroc \approx 0.999$ and $\auprc \approx 0.999$ on \texttt{gaussian state noise}. Conversely, at the challenging end of the spectrum, the metrics cluster around the balanced chance baseline: \texttt{lyapunov divergence} records an aggregate matched-time $\auroc = 0.539$ and $\auprc = 0.530$, while \texttt{cusum} yields $\auroc = 0.495$ and $\auprc = 0.556$. The aggregate matched-time AUPRC range (0.530--0.851) is thus valuable as a positive-class precision diagnostic, avoiding inflation caused by the extended post-onset interval.

The practical implication is that matched-time AUROC should serve as the principal threshold-free metric for discrimination on this benchmark, with matched-time AUPRC acting as a secondary diagnostic for positive-class precision. We continue to report raw trajectory metrics, null baselines, operating-point error, and latency as they address distinct questions: whether a detector distinguishes same-time clean and anomalous scores, whether it exploits temporal score drift, whether an operating threshold induces false alarms, and whether an alarm arrives in sufficient time to be useful.

\section{Discussion}
\label{sec:discussion}
The core contribution lies with the benchmark framework itself, rather than the specific ranking of detection methods. The detector and anomaly tables presented in Section~\ref{sec:results} are best viewed as a worked example of the framework's output within a single environment. Consequently, this discussion prioritises the framework's architecture, its extensibility, and the ease of integration evidenced by the LunarLander implementation, before addressing the conclusions drawn from the LunarLander results.

\subsection{Framework architecture and contracts}
\label{sec:framework-arch}
OOD-RL-Bench is organised as a compact Python package (\texttt{ood\_rl\_bench}), where distinct modules manage specific functions within the evaluation pipeline. The \texttt{core} module establishes the trajectory data structure and defines the abstract base classes \texttt{BaseDetector} and \texttt{BaseAnomalyInjector}. Concrete implementations for these roles are provided by the \texttt{anomalies}, \texttt{detectors}, \texttt{metrics}, and \texttt{environments} modules. The \texttt{scenarios} module offers end-to-end drivers that combine a policy, environment, injector set, and detector set, while the \texttt{utils} module supplies logging, deterministic seeding, and provenance recording (\texttt{run\_meta.yaml}). The orchestration layer is intentionally minimal and follows the data flow introduced in Figure~\ref{fig:framework-design}: scenario input is used to select the calibration, anomaly-execution, detector-scoring, metric, and reporting stages (Figure~\ref{fig:framework-design}A--H). Figure~\ref{fig:framework-architecture} summarises how the internal modules fit together. All interactions between components adhere to three compact data structures:

\begin{figure}[pos=htbp]
\centering
\includegraphics[width=\linewidth]{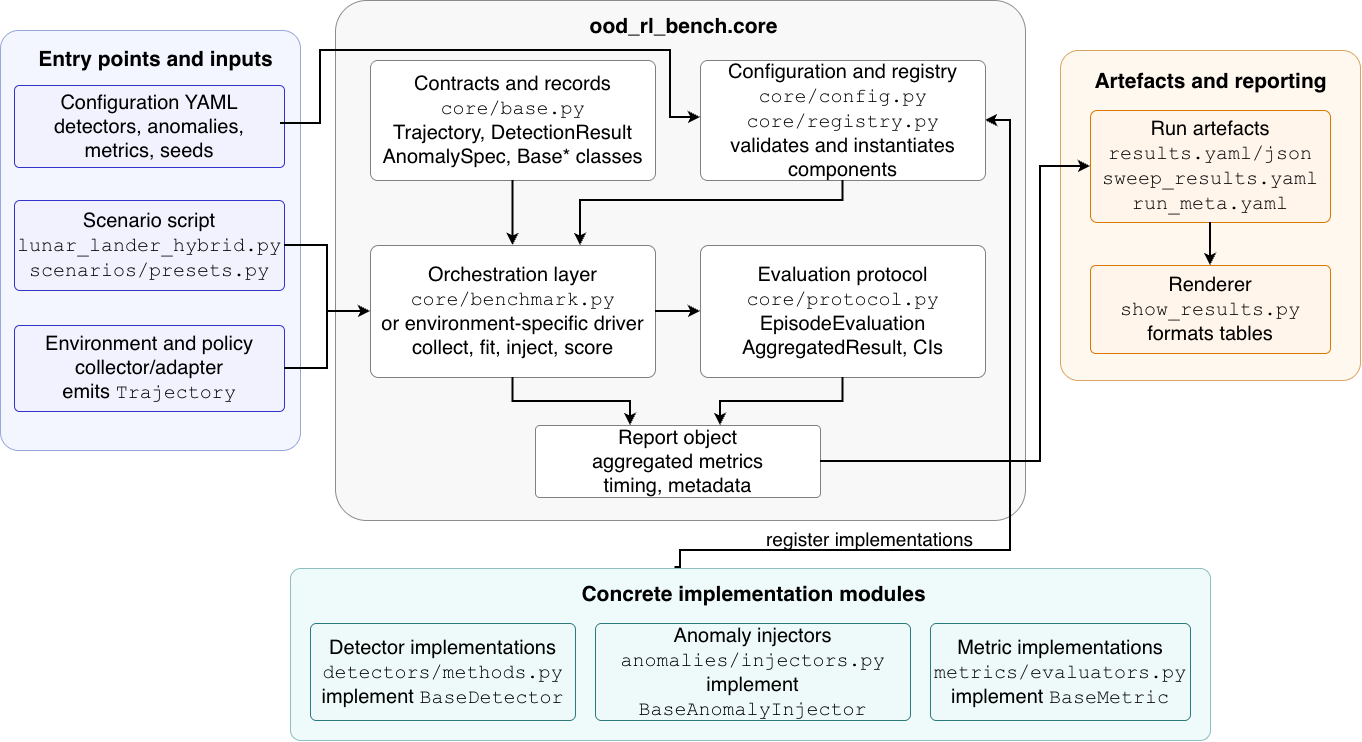}
\caption{OOD-RL-Bench internal architecture. Configuration and scenario entry points utilise the registry to select components, while core contracts delineate the data flow between modules; concrete detector, injector, and metric implementations register against these specifications, enabling orchestration and evaluation tiers to generate run artefacts for reporting.}
\label{fig:framework-architecture}
\end{figure}
\FloatBarrier

\begin{itemize}
\item \emph{Trajectory.} A fixed-length record exposing \texttt{states}, \texttt{actions}, \texttt{rewards}, \texttt{next\_states}, \texttt{dones}, and an \texttt{info} dictionary for environment-specific metadata. Every detector functions upon this structure; live-environment interaction is restricted to the trajectory collection phase for online injectors.
\item \emph{AnomalySpec.} Comprising a name, a severity scalar, an onset (and optional offset) step, and an injector-specific keyword dictionary. The specification is logged alongside the corrupted trajectory and, in conjunction with the seed and execution mode, identifies the perturbation responsible for the outcome.
\item \emph{DetectorOutput.} A score sequence of length $T$ alongside an optional thresholded mask; every metric within \texttt{metrics} processes this output in tandem with a ground-truth label vector.
\end{itemize}

Because scoring operates at the trajectory record level (Figure~\ref{fig:framework-design}D) rather than on the environment, the detector interface (Figure~\ref{fig:framework-design}E) does not require simulator access. Post-hoc injectors similarly operate directly upon the trajectory record (Figure~\ref{fig:framework-design}C1). Online injectors require only a minor environment-specific collector that applies the fault during rollout and subsequently emits the standard \texttt{Trajectory} structure (Figure~\ref{fig:framework-design}C2). This separation ensures uniform detector comparison while permitting dynamics faults to modify the policy's actual state visitation distribution.

\subsection{Extensibility}
\label{sec:framework-ext}
Extensibility is a central design requirement of OOD-RL-Bench. The framework was engineered to ensure that the three principal extension surfaces—detectors, injectors, and environments—remain independent and additive. Introducing a component along any of these surfaces does not require modification of the framework core: a new detector changes only the detector registry/configuration and a new anomaly injector changes only the injector implementation and its configuration block; the orchestration, metric aggregation, and reporting are unchanged.

\paragraph{Adding a detector.} A new detector subclasses \texttt{BaseDetector} and implements the detector methods shown in Figure~\ref{fig:framework-design}E: \texttt{fit(trajectories)}, invoked once on the in-distribution fit set, and \texttt{score(trajectory)} returning a score array of shape $(T,)$, invoked per evaluation episode. The detector determines internally which channels of the trajectory it utilises—states alone, $(s_t, a_t, s_{t+1})$ transitions, rewards, or any combination thereof—and the framework imposes no further constraint on representation, training procedure, or hardware backend. Registration consists of a single entry appended to the configuration YAML (Yet Another Markup Language) file's \texttt{detectors} list; no metric, runner, or scenario code is touched.

\paragraph{Adding an anomaly injector.} A post-hoc injector subclasses \texttt{BaseAnomalyInjector} and implements \texttt{inject(trajectory, onset\_step, offset\_step=None)}, returning the corrupted trajectory together with the binary ground-truth label vector through the post-hoc path in Figure~\ref{fig:framework-design}C1. Injector-specific parameters (e.g., \texttt{ramp\_steps}, \texttt{target\_action}, \texttt{max\_delay}) are declared as keyword arguments on the constructor and surfaced through the \texttt{anomaly\_configs\_single} and \texttt{anomaly\_severity\_sweep} blocks of the configuration YAML file. An online injector instead extends the environment adapter so that the perturbation is applied during rollout (Figure~\ref{fig:framework-design}C2), but it still returns the same trajectory-plus-label pair. The surrounding machinery—iteration over episodes, scoring of detectors, aggregation of metrics, recording of provenance—is reused unchanged.

\paragraph{Adding a metric.} A novel metric is a callable consuming a score sequence and a label vector, registered through the configuration YAML file's \texttt{metrics} block. The metric aggregation layer (Figure~\ref{fig:framework-design}G) is itself metric-agnostic: per-episode values are aggregated to mean and standard deviation, and the rendered output (Figure~\ref{fig:framework-design}H; Section~\ref{sec:framework-reporting}) inherits the additional metric without requiring further modification of the code. AUPRC, $\fpr_{0.95}$, segment-level precision and recall, and detection delay were each added through this single surface.

\paragraph{Adding an environment.} The framework treats an environment as a function that accepts a trained policy and produces a \texttt{Trajectory}. No shared environment interface beyond this contract is required: a Gymnasium environment, a custom physics simulator, or a logged interaction record are all acceptable trajectory sources, provided that the resultant trajectory exposes the six standard fields. The LunarLander integration includes the concrete environment adapter shipped with this release; the same trajectory contract can be used to implement additional Gymnasium or simulator adapters.

\subsection{Bringing up LunarLander: a worked example}
\label{sec:framework-lunar}
LunarLander-v3 acts as the definitive example for deploying OOD-RL-Bench to a new environment. The integration involves exactly four artefacts, none of which necessitate modifying the \texttt{ood\_rl\_bench} package itself:

\begin{enumerate}
\item \emph{Trained policy checkpoint.} A NoisyNet DQN model operating on the eight-dimensional flat state, trained via a standard script external to the framework. The framework places no restrictions on the architecture, training procedure, or training environment; it merely ingests the resultant \texttt{torch} \texttt{state\_dict} via the policy loader detailed below. This checkpoint is loaded once and reused for both in-distribution and out-of-distribution rollouts (Section~\ref{sec:setup}).
\item \emph{Policy loader and rollout functions.} The scenario file provides \texttt{\_load\_policy}, which instantiates the value network and loads the checkpoint, alongside rollout functions that emit \texttt{Trajectory} records. The clean/post-hoc path utilises \texttt{collect\_lunar\_trajectory} with the policy network, device, step count, seed, and policy mode as inputs. The online path applies actuator or dynamics faults during LunarLander-v3 interactions and records the commanded action, executed action, adjusted state transition, reward, and termination flag. These rollout functions represent the sole environment-aware code within the integration.
\item \emph{Configuration YAML.} A flat file (\texttt{lunar\_lander.yaml}) listing environment identifiers, training hyperparameters consumed by the policy script, benchmark parameters ($N_{\mathrm{ID}}$, $N_{\mathrm{OOD}}$, $T$), the detector list, and the per-injector default severity and severity-sweep grid. Switching the detector or injector sets requires only the modification of this file.
\item \emph{Scenario driver.} A top-level Python file (\texttt{lunar\_lander.py}) that invokes the loader, fits the detector suite once, routes each anomaly to the post-hoc or online collector, and writes the results alongside the \texttt{run\_meta.yaml} provenance record.
\end{enumerate}

The total environment-specific code remains compact: clean rollout, policy loading, and LunarLander-specific online perturbation logic reside outside the detector implementations, while the remaining infrastructure—eight detectors, nine anomaly injectors, eight metrics, deterministic seeding, matched-time result rendering, null-baseline diagnostics, and severity sweeping—is inherited from the package. A new environment with a comparable action and observation space is therefore expected to require only an adapter for clean rollout and, if online dynamics anomalies are needed, a small wrapper for applying faults, rather than a complete benchmark rewrite.

\subsection{Reproducibility and reporting}
\label{sec:framework-reporting}
Each execution generates three artefacts within the \texttt{output/<timestamp>\_<mode>/} directory: \texttt{results.yaml} (containing flat per-condition records), \texttt{sweep\_results.yaml} (containing nested per-injector severity grids where applicable), and \texttt{run\_meta.yaml} (detailing the framework version, Python and dependency versions, device, wall time, and the fully resolved configuration). The runner logs the execution mode for every anomaly and generates both post-hoc and online null baselines. The results schema encompasses raw metrics, matched-time metrics, null-derived diagnostics, and all-zero clean alarm rates. A symbolic link at \texttt{output/latest\_hybrid} consistently points to the most recent hybrid execution for easy inspection. The results files are intentionally serialised as human-readable YAML rather than a binary format, allowing them to be diffed, version-controlled, and post-processed without the need for external library dependencies. The accompanying \texttt{show\_results.py} renderer formats the matched-time per-anomaly and aggregate tables in the layout adopted throughout this paper, and transparently accepts both single-severity and sweep files. Collectively, these attributes render the framework's output trivially auditable: every tabulated value presented in Section~\ref{sec:results} corresponds to a single entry within a YAML file released alongside this manuscript, identified by detector name, anomaly name, execution mode, severity, metric, seed, and software versions used to generate the data.

\subsection{Mixed post-hoc and online execution: strengths and caveats}
The mixed protocol highlights a deliberate distinction between anomalies that can be faithfully captured in recorded data and those whose effects rely on closed-loop interaction. Post-hoc injection remains the most tractable approach for observation, temporal, and distributional perturbations: as every detector is presented with the identical perturbed trajectory, differences in metric values cannot be attributed to divergent control responses. Online collection is essential for action and dynamics faults because post-hoc modification would otherwise leave the state visitation distribution inconsistent with the action actually executed within the environment. The caveat lies in the fact that the online anomalies serve as LunarLander-specific proxy interventions: while they incorporate the policy's response to the fault during trajectory collection, the detectors are evaluated only after the trajectory has been gathered. These results should therefore be interpreted as mixed-mode trajectory-detection outcomes rather than constituting a complete real-time fault-response system.

\subsection{Results obtained on LunarLander}
\label{sec:results-discussion}
Three distinct structural patterns emerge, each of which the framework was explicitly designed to capture. Firstly, anomalies that displace the observation distribution or distort observable temporal structure are recovered with relative ease, encompassing both the post-hoc observation perturbations and \texttt{regime switch}. Secondly, matched-time negatives clarify action- and dynamics-related rows: \texttt{kde density} leads the \texttt{stuck actuator} row because the live fault alters the downstream state distribution, whereas \texttt{pedm} leads \texttt{dynamics scaling}, \texttt{action conditioned dynamics}, and \texttt{gradual drift}. Thirdly, the more difficult rows necessitate multiple diagnostics: \texttt{observation delay} and \texttt{action conditioned dynamics} demonstrate why matched-time AUROC, operating-point false-positive rate, and detection delay cannot be reduced to a single scalar ranking. These observations represent precisely the type of conclusion the framework is designed to facilitate; they emerge from the metric set, injector taxonomy, execution-mode split, and null-baseline, operating-point, and latency diagnostics rather than the design of any individual detector.

The \texttt{stuck actuator} result should not be interpreted as evidence that a marginal state-density detector identifies actuator faults causally. Within this implementation, the primary action field records the action commanded by the policy, while the action actually executed by the environment is retained as trajectory metadata. \texttt{pedm} therefore conditions on the commanded action and detects the fault solely through transition error; it does not directly score the mismatch between commanded and executed actions. \texttt{kde density}, by contrast, benefits because the live actuator fault rapidly displaces the downstream state distribution. The row therefore measures how detectable the downstream trajectory changes caused by the actuator fault are, rather than providing a direct diagnosis of the actuator mechanism itself.

\subsection{Limitations}
Several constraints impact both the framework and the interpretation of the reported results. (i) \emph{Single environment in the reported numbers}: The numerical results presented in Section~\ref{sec:results} originate solely from LunarLander-v3, utilising a single trained policy. While the framework's trajectory interface is designed to be environment-agnostic, adapting the approach for other environments—specifically those that are higher-dimensional, partially observed, or image-based—requires further implementation and empirical validation. (ii) \emph{Environment-specific online perturbation implementations}: Although the online actuator and dynamics fault families are conceptually portable across environments, precise configuration of concrete simulator hooks, affected control dimensions, and severity mappings is required prior to cross-environment comparisons. (iii) \emph{Matched-time scope}: Matched-time negatives provide a rigorous foundation for benchmark discrimination by contrasting anomalous scores with their clean counterparts at the corresponding timestep, thereby preventing detectors from capitalising on natural early-to-late score drift. It is important to recognise that this constitutes an evaluation protocol rather than a deployable detector; practical implementation therefore requires further threshold calibration, delay analysis, and false-alarm monitoring.

\section{Conclusion}
\label{sec:conclusion}
This manuscript introduces OOD-RL-Bench, a systematic benchmark for out-of-distribution detection applied to reinforcement-learning trajectories. The framework integrates eight evaluated detectors with nine anomaly mechanisms under a deterministic and reproducible mixed protocol: observation, temporal, and distributional anomalies are evaluated post hoc, while action and dynamics anomalies are evaluated through online LunarLander rollouts before being scored by the same detector suite. Initial evaluations on LunarLander establish a nuanced hierarchy for anomaly difficulty and detector performance under matched-time negatives. Specifically, observation corruptions and regime switches present straightforward challenges, stuck-actuator detection is strongest for a state-density signal in the current suite, and online dynamics rows are strongest for \texttt{pedm} but expose a latency trade-off. Observation delay and action-conditioned dynamics require careful interpretation through null-baseline, false-positive-rate, and latency diagnostics. To facilitate subsequent methodological advancements against a unified baseline, we release the framework, trained policy, and comprehensive results at \url{https://github.com/ood-rl-bench/ood-rl-bench}.

\section*{Acknowledgements}
This research was supported by research funding from Deakin University.

\bibliographystyle{cas-model2-names}
\bibliography{references}

@book{sutton2018,
  author    = {Sutton, Richard S. and Barto, Andrew G.},
  title     = {Reinforcement Learning: An Introduction},
  edition   = {2},
  publisher = {MIT Press},
  year      = {2018}
}

@article{mnih2015,
  author  = {Mnih, Volodymyr and Kavukcuoglu, Koray and Silver, David and Rusu, Andrei A. and Veness, Joel and Bellemare, Marc G. and Graves, Alex and Riedmiller, Martin and Fidjeland, Andreas K. and Ostrovski, Georg and Petersen, Stig and Beattie, Charles and Sadik, Amir and Antonoglou, Ioannis and King, Helen and Kumaran, Dharshan and Wierstra, Daan and Legg, Shane and Hassabis, Demis},
  title   = {Human-level control through deep reinforcement learning},
  journal = {Nature},
  volume  = {518},
  number  = {7540},
  pages   = {529--533},
  year    = {2015},
  doi     = {10.1038/nature14236}
}

@article{brockman2016,
  author  = {Brockman, Greg and Cheung, Vicki and Pettersson, Ludwig and Schneider, Jonas and Schulman, John and Tang, Jie and Zaremba, Wojciech},
  title   = {OpenAI Gym},
  journal = {arXiv preprint arXiv:1606.01540},
  year    = {2016},
  doi     = {10.48550/arXiv.1606.01540}
}

@article{towers2024,
  author  = {Towers, Mark and Kwiatkowski, Ariel and Terry, Jordan K. and Balis, John U. and De Cola, Gianluca and Deleu, Tristan and Goul{\~a}o, Manuel and Kallinteris, Andreas and Krimmel, Markus and KG, Arjun and Perez-Vicente, Rodrigo and Pierr{\'e}, Andrea and Schulhoff, Sander and Tai, Jun Jet and Tan, Hannah and Younis, Omar G.},
  title   = {Gymnasium: A Standard Interface for Reinforcement Learning Environments},
  journal = {arXiv preprint arXiv:2407.17032},
  year    = {2024},
  doi     = {10.48550/arXiv.2407.17032}
}

@inproceedings{hendrycks2017,
  author    = {Hendrycks, Dan and Gimpel, Kevin},
  title     = {A baseline for detecting misclassified and out-of-distribution examples in neural networks},
  booktitle = {International Conference on Learning Representations},
  year      = {2017}
}

@inproceedings{liang2018,
  author    = {Liang, Shiyu and Li, Yixuan and Srikant, R.},
  title     = {Enhancing the Reliability of Out-of-distribution Image Detection in Neural Networks},
  booktitle = {International Conference on Learning Representations},
  year      = {2018}
}

@inproceedings{lee2018,
  author    = {Lee, Kimin and Lee, Kibok and Lee, Honglak and Shin, Jinwoo},
  title     = {A Simple Unified Framework for Detecting Out-of-Distribution Samples and Adversarial Attacks},
  booktitle = {Advances in Neural Information Processing Systems},
  volume    = {31},
  pages     = {7167--7177},
  year      = {2018}
}

@inproceedings{lakshminarayanan2017,
  author    = {Lakshminarayanan, Balaji and Pritzel, Alexander and Blundell, Charles},
  title     = {Simple and Scalable Predictive Uncertainty Estimation Using Deep Ensembles},
  booktitle = {Advances in Neural Information Processing Systems},
  volume    = {30},
  pages     = {6405--6416},
  year      = {2017}
}

@inproceedings{yang2022,
  author    = {Yang, Jingkang and Wang, Pengyun and Zou, Dejian and Zhou, Zitang and Ding, Kunyuan and Peng, Wenxuan and Wang, Haoqi and Chen, Guangyao and Li, Bo and Sun, Yiyou and Du, Xuefeng and Zhou, Kaiyang and Zhang, Wayne and Hendrycks, Dan and Li, Yixuan and Liu, Ziwei},
  title     = {{OpenOOD}: Benchmarking Generalized Out-of-Distribution Detection},
  booktitle = {Advances in Neural Information Processing Systems},
  volume    = {35},
  pages     = {32598--32611},
  year      = {2022}
}

@article{zhang2023,
  author        = {Zhang, Jingyang and Yang, Jingkang and Wang, Pengyun and Wang, Haoqi and Lin, Yueqian and Zhang, Haoran and Sun, Yiyou and Du, Xuefeng and Li, Yixuan and Liu, Ziwei and Chen, Yiran and Li, Hai},
  title         = {{OpenOOD} v1.5: Enhanced benchmark for out-of-distribution detection},
  journal       = {arXiv preprint arXiv:2306.09301},
  year          = {2023},
  doi           = {10.48550/arXiv.2306.09301}
}

@article{paparrizos2022,
  author        = {Paparrizos, John and Kang, Yuhao and Boniol, Paul and Tsay, Ruey S. and Palpanas, Themis and Franklin, Michael J.},
  title         = {{TSB-UAD}: An end-to-end benchmark suite for univariate time-series anomaly detection},
  journal       = {Proceedings of the VLDB Endowment},
  volume        = {15},
  number        = {8},
  pages         = {1697--1711},
  year          = {2022},
  doi           = {10.14778/3529337.3529354}
}

@article{qiu2025,
  author  = {Qiu, Xianfei and Qiu, Wanghui and Hu, Shiyan and Zhou, Lekui and Wu, Xingjian and Zhengyu, Li and Guo, Chenjuan and Zhou, Aoying and Sheng, Zhenli},
  title   = {{TAB}: Unified benchmarking of time series anomaly detection methods},
  journal = {Proceedings of the VLDB Endowment},
  volume  = {18},
  number  = {9},
  pages   = {2775-2789},
  year    = {2025},
  doi     = {10.14778/3746405.3746407}
}

@article{mohammed2021,
  author        = {Mohammed, Aaqib Parvez and Valdenegro-Toro, Matias},
  title         = {Benchmark for out-of-distribution detection in deep reinforcement learning},
  journal       = {arXiv preprint arXiv:2112.02694},
  year          = {2021},
  doi           = {10.48550/arXiv.2112.02694}
}

@article{danesh2021,
  author        = {Danesh, Mohamad H. and Fern, Alan},
  title         = {Out-of-distribution dynamics detection: {RL}-relevant benchmarks and results},
  journal       = {arXiv preprint arXiv:2107.04982},
  year          = {2021},
  doi           = {10.48550/arXiv.2107.04982}
}

@inproceedings{mueller2022,
  author    = {M{\"u}ller, Robert and Illium, Steffen and Phan, Thomy and Haider, Tom and Linnhoff-Popien, Claudia},
  title     = {Towards anomaly detection in reinforcement learning},
  booktitle = {BlueSky Ideas of the 21st International Conference on Autonomous Agents and Multiagent Systems},
  pages     = {1799--1803},
  publisher = {International Foundation for Autonomous Agents and Multiagent Systems},
  year      = {2022}
}

@inproceedings{haider2023,
  author    = {Haider, Tom and Roscher, Karsten and Schmoeller da Roza, Felippe and G{\"u}nnemann, Stephan},
  title     = {Out-of-distribution detection for reinforcement learning agents with probabilistic dynamics models},
  booktitle = {Proceedings of the 2023 International Conference on Autonomous Agents and Multiagent Systems},
  pages     = {851--859},
  publisher = {International Foundation for Autonomous Agents and Multiagent Systems},
  year      = {2023}
}

@inproceedings{nasvytis2024,
  author    = {Nasvytis, Linas and Sandbrink, Kai and Foerster, Jakob and Franzmeyer, Tim and {Schroeder de Witt}, Christian},
  title     = {Rethinking Out-of-Distribution Detection for Reinforcement Learning: Advancing Methods for Evaluation and Detection},
  booktitle = {Proceedings of the 23rd International Conference on Autonomous Agents and Multiagent Systems},
  pages     = {1445--1453},
  publisher = {International Foundation for Autonomous Agents and Multiagent Systems},
  year      = {2024}
}

@article{prashant2025,
  author  = {Prashant, Mohit and Easwaran, Arvind and Das, Suman and Yuhas, Michael},
  title   = {Guaranteeing Out-Of-Distribution Detection in Deep {RL} via Transition Estimation},
  journal = {Proceedings of the AAAI Conference on Artificial Intelligence},
  volume  = {39},
  number  = {12},
  pages   = {12452--12460},
  year    = {2025},
  doi     = {10.1609/aaai.v39i12.33357}
}

@inproceedings{chua2018,
  author    = {Chua, Kurtland and Calandra, Roberto and McAllister, Rowan and Levine, Sergey},
  title     = {Deep Reinforcement Learning in a Handful of Trials Using Probabilistic Dynamics Models},
  booktitle = {Advances in Neural Information Processing Systems},
  volume    = {31},
  pages     = {4759--4770},
  year      = {2018}
}

@article{mahalanobis1936,
  author  = {Mahalanobis, Prasanta Chandra},
  title   = {On the generalised distance in statistics},
  journal = {Proceedings of the National Institute of Sciences of India},
  volume  = {2},
  pages   = {49--55},
  year    = {1936}
}

@inproceedings{sakurada2014,
  author    = {Sakurada, Mayu and Yairi, Takehisa},
  title     = {Anomaly detection using autoencoders with nonlinear dimensionality reduction},
  booktitle = {Proceedings of the {MLSDA} 2014 2nd Workshop on Machine Learning for Sensory Data Analysis},
  series    = {{MLSDA}'14},
  pages     = {4--11},
  publisher = {Association for Computing Machinery},
  year      = {2014},
  doi       = {10.1145/2689746.2689747}
}

@article{parzen1962,
  author  = {Parzen, Emanuel},
  title   = {On estimation of a probability density function and mode},
  journal = {The Annals of Mathematical Statistics},
  volume  = {33},
  number  = {3},
  pages   = {1065--1076},
  year    = {1962},
  doi     = {10.1214/aoms/1177704472}
}

@article{page1954,
  author  = {Page, E. S.},
  title   = {Continuous inspection schemes},
  journal = {Biometrika},
  volume  = {41},
  number  = {1--2},
  pages   = {100--115},
  year    = {1954},
  doi     = {10.1093/biomet/41.1-2.100}
}

@book{basseville1993,
  author    = {Basseville, Michele and Nikiforov, Igor V.},
  title     = {Detection of Abrupt Changes: Theory and Application},
  publisher = {Prentice Hall},
  address   = {Englewood Cliffs, NJ},
  year      = {1993}
}

@article{fulcher2018,
  author  = {Fulcher, Ben D. and Jones, Nick S.},
  title   = {{hctsa}: A Computational Framework for Automated Time-Series Phenotyping Using Massive Feature Extraction},
  journal = {Cell Systems},
  volume  = {5},
  number  = {5},
  pages   = {527--531.e3},
  year    = {2017},
  doi     = {10.1016/j.cels.2017.10.001}
}

@article{wolf1985,
  author  = {Wolf, Alan and Swift, Jack B. and Swinney, Harry L. and Vastano, John A.},
  title   = {Determining Lyapunov exponents from a time series},
  journal = {Physica D: Nonlinear Phenomena},
  volume  = {16},
  number  = {3},
  pages   = {285--317},
  year    = {1985},
  doi     = {10.1016/0167-2789(85)90011-9}
}

@book{kantz2004,
  author    = {Kantz, Holger and Schreiber, Thomas},
  title     = {Nonlinear Time Series Analysis},
  edition   = {2},
  publisher = {Cambridge University Press},
  year      = {2004}
}

@article{fawcett2006,
  author  = {Fawcett, Tom},
  title   = {An introduction to {ROC} analysis},
  journal = {Pattern Recognition Letters},
  volume  = {27},
  number  = {8},
  pages   = {861--874},
  year    = {2006},
  doi     = {10.1016/j.patrec.2005.10.010}
}

@inproceedings{davis2006,
  author    = {Davis, Jesse and Goadrich, Mark},
  title     = {The relationship between precision-recall and {ROC} curves},
  booktitle = {Proceedings of the 23rd International Conference on Machine Learning},
  pages     = {233--240},
  publisher = {Association for Computing Machinery},
  year      = {2006},
  doi       = {10.1145/1143844.1143874}
}

\end{document}